\documentclass[12pt]{article}
\usepackage[a4paper,margin=18mm]{geometry}
\usepackage{amsmath,amssymb,booktabs,graphicx,tabularx,float}

\usepackage[super,comma,sort&compress]{natbib}

\usepackage{xr-hyper}
\usepackage[hidelinks]{hyperref}
\usepackage{microtype,newtxtext,newtxmath,placeins,xurl,setspace}
\renewcommand{\harvardurl}[1]{\url{#1}}
\usepackage[font=small,labelfont=bf,labelsep=period]{caption}
\hypersetup{pdftitle={Segment Anything Pancreas: A unified MRI segmentation model for pancreatic cancer radiotherapy}}


\hypersetup{pdfauthor={Zach Eidex, Mojtaba Safari, Kundan Thind, Parag J. Parikh, Zhen Tian, Xiaofeng Yang}}

\ifdefined\PMBArticleOnly
\newcommand{\suppref}[1]{\ref*{#1}}
\else
\newcommand{\suppref}[1]{\ref{#1}}
\fi
\begin{document}
\begin{center}
{\LARGE\bfseries Segment Anything Pancreas: A Unified MRI Segmentation Model for Pancreatic Cancer Radiotherapy\par}
\vspace{9pt}
{\normalsize Zach Eidex\textsuperscript{1}, Mojtaba Safari\textsuperscript{2}, Kundan Thind\textsuperscript{3},
Parag J. Parikh\textsuperscript{3},\\[3pt]
Zhen Tian\textsuperscript{2,*} and Xiaofeng Yang\textsuperscript{1,2,*}\par}
\vspace{7pt}
{\small
\textsuperscript{1}Department of  Computer Science and Radiation Oncology, Emory University, Atlanta, GA, USA\\
\textsuperscript{2}Department of Radiation and Cellular Oncology, University of Chicago, Chicago, IL 60637, USA\\
\textsuperscript{3}Department of Radiation Oncology, Henry Ford Health, Detroit, MI, USA\par}
\vspace{5pt}
{\small{*}Email: \href{mailto:ztian@bsd.uchicago.edu (ZT); xfyang@uchicago.edu (XY)}{ztian@bsd.uchicago.edu (ZT) and xfyang@uchicago.edu (XY)}\par}
\end{center}
\vspace{5pt}
\begin{center}\textbf{Abstract}\end{center}

\noindent\textbf{Background:}
Accurate delineation of pancreatic tumors and organs at risk (OARs) is essential for MRI-guided radiotherapy. However, limited tumor annotations and incomplete anatomical labels across MRI datasets hinder the development of unified segmentation models.

\noindent\textbf{Purpose:}
We propose Pancreas-MedNeXt, a unified model for segmenting the whole pancreas, pancreatic subregions, tumors, and surrounding OARs from a single MRI volume.

\noindent\textbf{Methods:}
Pancreas-MedNeXt uses a shared MedNeXt encoder-decoder with supervised whole-pancreas and subregion pretraining, followed by multitask fine-tuning. Three output heads jointly predict the pancreas and its subregions (head, body, tail), the tumor, and the OARs (stomach, duodenum, bilateral kidneys). An annotation-aware loss allows training on partially labeled data without treating unannotated structures as background. The model was developed on four public MRI datasets (PanSegData, Cyst-X, AMOS, PANTHER). Pancreas and tumor segmentation were evaluated by three-fold cross-validation on 142 PANTHER patients (92 diagnostic MRI, 50 MR-Linac MRI). Subregion and OAR segmentation were evaluated on held-out Cyst-X and AMOS test sets, and pancreas and OAR segmentation on 33 external T1-weighted TotalSegmentator MRI scans. Performance was compared with nnU-Net, MedNeXt, VISTA3D, and Swin UNETR using the Dice similarity coefficient (DSC) and surface metrics, with 95\% confidence intervals (CIs) from paired bootstrap resampling.

\noindent\textbf{Results:}
Pancreas-MedNeXt achieved the highest mean pancreas and tumor DSC on both diagnostic and MR-Linac MRI among the evaluated methods. Tumor DSC was $58.41 \pm 26.53\%$ and $42.55 \pm 23.21\%$,  exceeding task-specific MedNeXt by 4.94 (95\% CI: 0.28--9.45) and 5.89 (95\% CI: 2.37--9.72) percentage points, respectively. Whole-pancreas DSC was $85.73 \pm 8.35\%$ and $70.66 \pm 11.28\%$. Supervised pretraining improved head, body, and tail DSC by 2.01, 3.00, and 3.78 percentage points over the same model trained from scratch. On AMOS, the model achieved the highest DSC for the pancreas, duodenum, and right kidney. On external MRI, it achieved the highest DSC for all five organs, exceeding the next-best method by 4.92–8.85 percentage points.

\noindent\textbf{Conclusion:}
Pancreas-MedNeXt segments tumors, pancreatic subregions, and OARs in a single model across diagnostic and MR-Linac MRI. This approach could streamline contouring and support quantitative image analysis and treatment planning for pancreatic cancer radiotherapy.

\medskip

\section{Introduction}
Pancreatic cancer remains one of the most lethal types of cancer, with five-year relative survival below 15\% in the United States \citep{seer_pancreas}. For patients with nonmetastatic, unresectable disease, radiotherapy provides a local treatment option after systemic therapy. Ablative and dose-escalated approaches have shown promising local control, motivating techniques that improve tumor coverage \citep{parikh_smart,reyngold_ablative}. MRI serves complementary roles in the treatment of pancreatic tumors. Diagnostic MRI characterizes the lesion and its anatomical extent, whereas a magnetic resonance linear accelerator (MR-Linac) acquires images in the treatment position to guide radiation delivery and online plan adaptation. Treatment planning requires delineation of the tumor and surrounding organs at risk (OARs), particularly the stomach and duodenum, whose proximity often limits the prescribed dose \citep{grimbergen_consensus}. Identifying both the tumor and the involved pancreatic subregion is important for radiation therapy contouring based on the NRG Oncology consensus guidelines used in the ongoing LAP100 study (NCT06958328) \citep{sanford2025atlas}. Gastrointestinal OARs undergo substantial interfraction changes in position and shape, motivating contour reassessment on the images acquired for each treatment fraction \citep{alam_gi_motion}. Repeated manual contour review and correction are time-consuming components of online adaptive radiotherapy \citep{shojaei}. Automated segmentation could assist this process by delineating the lesion, gland and adjacent normal tissues, especially if the approach can cope with visual differences between MR-Linac and diagnostic MRI \citep{eidex_mrgrt}. A unified model that delineates the pancreatic tumor, gland and adjacent OARs across diagnostic and treatment MRI could support anatomical assessment and contouring throughout the radiotherapy workflow.

Automated segmentation has been investigated for pancreatic tumor assessment and MRI-guided radiotherapy. \citet{liang2020} investigated pancreatic tumor segmentation on multiparametric MRI, and \citet{shojaei} developed abdominal OAR auto-contouring for adaptive MRI-guided pancreatic radiotherapy. PanSegNet extended whole-pancreas segmentation across multicenter T1-weighted, T2-weighted and CT datasets, addressing the substantial variability in pancreatic appearance across imaging protocols \citep{pansegnet}. For more detailed gland assessment, \citet{unifiedpancreas} transferred whole-pancreas representations learned from CT and MRI to head, body and tail segmentation. Staged anatomical transfer also appears in PANTHER methods, which use broad anatomy and pancreatic lesion supervision before adapting to diagnostic and therapeutic MRI \citep{pantherwinner}. These studies establish a rationale for learning pancreatic anatomy before more specialized segmentation tasks. More broadly, DoDNet demonstrated organ and tumor segmentation from partially labeled datasets, while TotalSegmentator MRI provides extensive anatomical coverage across MRI acquisitions \citep{dodnet,totalsegmri}. VISTA3D and the MRI foundation model Triad further demonstrate the use of pretrained representations across segmentation tasks \citep{vista3d,nvctmr,triad_yang}. Together, this literature supports transfer learning and shared representations, but the cited pancreatic studies evaluate gland, tumor, regional anatomy or OAR segmentation in separate task configurations. Whether a single MRI model can retain regional pancreatic anatomy while learning tumors and surrounding organs remains a clinically relevant question, particularly for MR-Linac imaging, where a model must delineate the lesion within the anatomy available for treatment guidance.

Pancreatic tumor, glandular anatomy and OAR segmentation serve distinct purposes in pancreatic cancer care. Tumor masks describe the lesion, while OAR masks support assessment of nearby normal tissue during radiotherapy planning. Whole-pancreas masks provide an anatomical reference for tumor localization and assessment of gland involvement, while head, body and tail masks permit regional measurements that would be obscured by a whole-gland average. Pancreatic subregions have been used for regional fat analysis, illustrating their role in quantitative imaging beyond the definition of a treatment target \citep{pancreasparts}. A model providing all of these masks could support tumor assessment, organ contouring and regional imaging within the same examination. Training such a model is complicated by incomplete annotation coverage. PanSegData and Cyst-X provide whole-pancreas labels, the Cyst-X subregion release supplies head, body and tail masks, PANTHER provides pancreas and tumor labels, and AMOS supplies abdominal organ annotations \citep{pansegnet,cystx,unifiedpancreas,panther,amos}. No single source used here provides the complete set. Simply pooling their label maps would assign contradictory meanings to background, because a visible structure may be annotated in one dataset and unannotated in another. The labels also describe different anatomical levels. Whole-pancreas masks overlap the head, body and tail, and a tumor may occupy part of a subregion or extend beyond the apparent gland. Multitask training must therefore use available supervision without treating missing annotations as negative examples, while preserving the relationship between whole-gland anatomy, regional anatomy and tumor extent. For MRI, this also permits scans acquired with different clinical protocols to contribute to a common anatomical representation. An anatomy-first curriculum is attractive because it uses the more widely available gland labels before introducing the less extensively annotated tumor task, while retaining the regional and organ outputs required for broader pancreatic assessment.

We propose Pancreas-MedNeXt, a unified MRI model that combines supervised pancreatic pretraining with an annotation-aware multitask objective (https://github.com/zacheidex/Segment-Anything-Pancreas). Its training curriculum progresses from whole-pancreas segmentation to subregion segmentation and then to simultaneous pancreatic, tumor and abdominal organ segmentation. The final model accepts one MRI volume and produces all nine region masks with a shared MedNeXt encoder-decoder. Whole-pancreas and subregion outputs share an anatomical hierarchy, while a separate tumor output permits overlap with the gland. This formulation allows datasets with incomplete annotations to supervise the same model without requiring additional manual labeling. The study addresses whether this approach improves pancreas and tumor segmentation across diagnostic and MR-Linac MRI while retaining regional and abdominal organ performance. The principal contributions are:
\begin{enumerate}
\item A unified MRI model for the whole pancreas, head, body, tail, tumor, stomach, duodenum, right kidney and left kidney. Evaluation demonstrates improved pancreas and tumor segmentation on diagnostic and MR-Linac MRI, together with subregion and abdominal organ capabilities on their respective datasets.
\item A supervised MRI training curriculum that transfers whole-pancreas and subregion representations into multitask fine-tuning, together with a matched from-scratch ablation evaluating the contribution of this pretraining strategy.
\item Compact subregion, tumor and OAR heads attached to one shared decoder, with an annotation-aware loss for learning from partially labeled datasets. The output design preserves the relationship between whole pancreas and its subregions while allowing the tumor mask to overlap pancreatic anatomy.
\end{enumerate}

\FloatBarrier
\section{Materials and methods}
\subsection{Datasets and preprocessing}\label{sec:datasets}
This retrospective study used four public diagnostic MRI datasets for model development: AMOS, PanSegData, Cyst-X and PANTHER. Whole-pancreas pretraining used AMOS, the PanSegData release associated with PanSegNet, and eligible additional Cyst-X scans \citep{amos,pansegnet,cystx}. PANTHER was reserved for the final multitask stage. The initial anatomical dataset contained 649 training and 55 validation scans from AMOS and PanSegData. Continued whole-pancreas pretraining incorporated Cyst-X, increasing these counts to 1,193 and 155 scans, respectively. Eligible AMOS and Cyst-X images subsequently supplied their more detailed labels for organ and subregion segmentation. Partitions were defined at the patient level, with multiple acquisitions from the same patient assigned to the same partition. Image-mask checks compared dimensions, orientation, origin and voxel spacing, and cases with unresolved alignment discrepancies were excluded. Table~\ref{tab:data} summarizes the datasets, partition sizes and annotated structures. Pretraining and fine-tuning counts describe their respective training stages rather than additional independent cohorts.

PANTHER comprises 142 annotated patients, including 92 diagnostic-MRI and 50 MR-Linac patients, each contributing one labeled volume \citep{panther,pantherdata}. The diagnostic images are arterial-phase contrast-enhanced T1-weighted acquisitions, whereas the therapeutic cohort contains T2-weighted MR-Linac planning MRI. Three-fold cross-validation rotated patient-level test sets of 48, 47 and 47 patients, stratified by diagnostic or MR-Linac acquisition, with 15 inner-validation patients per fold. The remaining 79, 80 and 80 patients supplied training images. Each patient was evaluated once using the model trained on the corresponding outer training partition. The first fold contained the original local patient holdout of 31 diagnostic and 17 MR-Linac patients. Reference labels distinguish pancreatic tissue and tumor, and their union defines the reported whole pancreas. Thus, pancreas evaluation includes tumor-bearing tissue, while tumor evaluation separately measures lesion delineation. The local patient holdout was inspected during model development, and the reported cross-validation constitutes exploratory evaluation of the selected configuration.

The Cyst-X head, body and tail reference masks were made available upon request and publicly released by the Machine and Hybrid Intelligence Lab \citet{unifiedpancreas}. Of 1,437 released masks, 129 were excluded because image-mask coordinate disagreements could not be resolved, and one was a within-release duplicate. The retained 1,307 scans were divided into 1,064 training, 142 validation and 101 test scans, representing 419, 59 and 43 patients. Training included 468 T1-weighted and 596 T2-weighted scans, and testing included 44 T1-weighted and 57 T2-weighted scans. Released head, body and tail labels were used directly, with their union defining the reference whole pancreas. The subregion release revises some boundaries relative to the earlier binary masks, so its labels were used consistently for both subregion training and evaluation. AMOS contributed 60 MRI patients, divided into 42 training, nine validation and nine test patients, with annotations for pancreas, stomach, duodenum and bilateral kidneys. Across the three multitask datasets, the complete training, validation and test collection contained 1,509 scans from 723 patients. Cyst-X and AMOS partitions remained fixed during PANTHER cross-validation, and their test results used the unified model from the first PANTHER fold. All PANTHER patients and the Cyst-X and AMOS test patients were excluded from both pretraining tasks.

External evaluation used 33 scans with T1-weighted contrast from the released TotalSegmentator MRI test partition \citep{totalsegmri}. These scans were identified by retrospective review of available acquisition metadata and images from 56 eligible scans. Scans with T2-weighted or indeterminate contrast were excluded from this sequence-restricted analysis. Organ coverage varied with the imaged anatomy, yielding references for 33 pancreata, 33 stomachs, 31 duodenums, 31 right kidneys and 33 left kidneys. Quantitative evaluation followed the structures annotated in each collection, even though the unified model generated its complete set of outputs for every input scan.

All images were preprocessed using the nnU-Net v2 pipeline \citep{nnunet2021}, with nonzero cropping, per-image MRI Z-score normalization and resampling to $3.0\times1.1875\times1.1875$ mm in array-axis order. Training used patches of $56\times160\times224$ voxels and a batch size of two. Spatial and intensity augmentation followed the shared nnU-Net configuration, including foreground-aware patch sampling. Each example contained one MRI volume in a single input channel. T1-weighted, T2-weighted, diagnostic and MR-Linac scans were processed independently, without paired sequences or intersequence registration. The single-channel input allowed patients with only one available sequence to contribute to the same model, without requiring both T1-weighted and T2-weighted acquisitions for every examination. Patches were padded internally to multiples of 16 for the MedNeXt encoder-decoder and cropped back before loss calculation or combination of inference windows. Predictions were resampled to the original image geometry for comparison with the native reference masks. This preserved physical distances in millimeters and avoided evaluating contours solely at the resampled training resolution.

\begin{table}[H]
\caption{Datasets, annotated structures, and MRI scan counts used for model development and external evaluation.}\label{tab:data}
\centering\small\setlength{\tabcolsep}{6pt}
\begin{tabularx}{\linewidth}{@{\hspace{4pt}}>{\raggedright\arraybackslash}p{39mm}>{\raggedright\arraybackslash}X>{\raggedright\arraybackslash}p{14mm}>{\raggedright\arraybackslash}p{20mm}>{\raggedright\arraybackslash}p{14mm}@{\hspace{6pt}}}
\toprule
 & \multicolumn{1}{c}{Structures} & \multicolumn{1}{c}{Train} & \multicolumn{1}{c}{Validation} & \multicolumn{1}{c@{\hspace{6pt}}}{Test} \\
\midrule
\multicolumn{5}{@{\hspace{4pt}}l}{\textit{Whole-pancreas pretraining}} \\
Cyst-X & Pancreas & 544 & 100 & - \\
PanSegData & Pancreas & 607 & 46 & - \\
AMOS & Pancreas & 42 & 9 & - \\
\midrule
\multicolumn{5}{@{\hspace{4pt}}l}{\textit{Unified training}} \\
Cyst-X & Pancreas, head, body, tail & 1,064 & 142 & 101 \\
\multicolumn{5}{@{\hspace{4pt}}l}{PANTHER} \\
Diagnostic MRI & Pancreas, tumor & 51 & 10 & 31 \\
MR-Linac MRI & Pancreas, tumor & 28 & 5 & 17 \\
AMOS & Pancreas, stomach, duodenum, right and left kidneys & 42 & 9 & 9 \\
\midrule
\multicolumn{5}{@{\hspace{4pt}}l}{\textit{External evaluation}} \\
TotalSegmentator MRI (T1-weighted) & Pancreas, stomach, duodenum, right and left kidneys & - & - & 33 \\
\bottomrule
\end{tabularx}
\end{table}

\subsection{Model architecture}
Pancreas-MedNeXt uses the original MedNeXt-M convolutional encoder-decoder with kernel size three, a base feature width of 32, residual blocks and multiresolution skip connections \citep{mednext}. The nine resolution stages use expansion ratios $[2,3,4,4,4,4,4,3,2]$ and block counts $[3,4,4,4,4,4,4,4,3]$. One encoder and one decoder serve all segmentation tasks. Three output heads are attached at each decoder supervision scale, each using a learned $1\times1\times1$ convolution to map the shared features to class logits. The subregion head predicts four classes: background, pancreatic body, head and tail. The OAR head predicts background and the four abdominal organs, while the tumor head predicts background and tumor. A separate softmax is applied within each head. The resulting 11 channels are converted to nine region probabilities: whole pancreas, body, head, tail, stomach, duodenum, right kidney, left kidney and tumor. The model contains approximately 17.56 million parameters. The additional output heads increase the parameter count by less than 0.1\% relative to the four-class MedNeXt configuration, retaining the same feature extraction and reconstruction network. Sharing decoder features allows each annotated task to update the representation used by the other outputs without maintaining separate task-specific decoders.

Let $\mathbf{p}(x)$ denote the softmax probabilities of the subregion head at voxel $x$, with $p_0$ representing anatomical background and $p_1,p_2,p_3$ representing body, head and tail. The whole-pancreas probability is
\begin{equation}
p_{\mathrm{pan}}(x)=p_1(x)+p_2(x)+p_3(x).
\end{equation}
A whole-pancreas reference therefore supervises the combined probability of all three regions without assigning an unsupported subregion identity. When a subregion annotation is available, it additionally supervises the corresponding anatomical component. If $q_j(x)$ denotes an OAR probability within the OAR-head softmax, its region probability is $p_0(x)q_j(x)$. The OAR head thus subdivides the background of the anatomical pancreas prediction. Tumor probability is obtained independently from the binary tumor head, permitting a lesion and a glandular subregion to occupy the same voxel. The model does not place tumor in competition with head, body and tail within one exclusive classifier. This distinction is necessary because anatomical location and tumor involvement are different attributes of the tissue. The probabilities are transformed into region logits for the binary region losses described below.

\begin{figure}[t]
\centering\includegraphics[width=\linewidth]{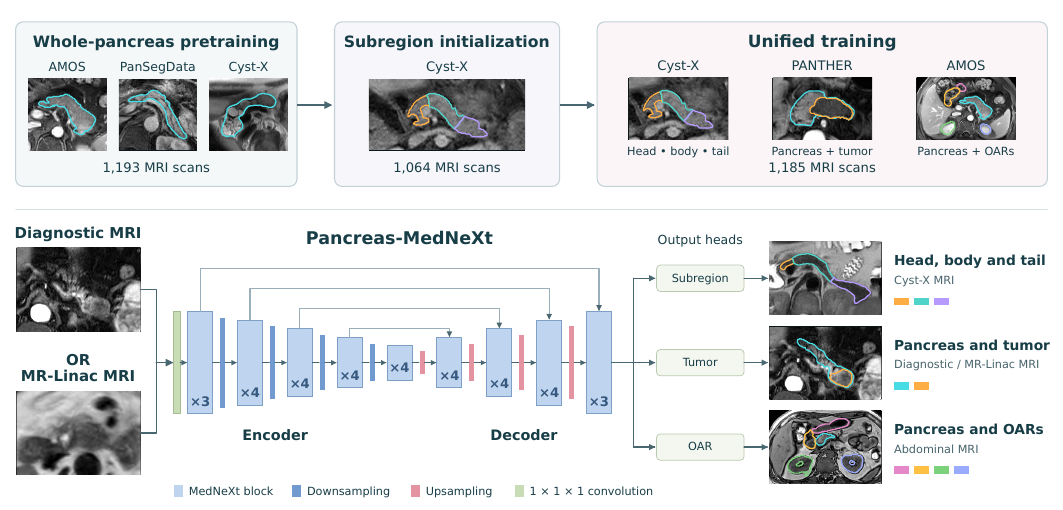}
\caption{Pancreas-MedNeXt training and segmentation workflow. Whole-pancreas and subregion pretraining initialize a shared encoder-decoder with subregion, tumor and OAR output heads. Diagnostic or MR-Linac MRI is processed through the same single-channel input. Examples show the final masks, with whole pancreas derived from subregion and tumor predictions.}\label{fig:method}
\end{figure}

At inference, the whole-pancreas mask is the union of voxels with $p_{\mathrm{pan}}>0.5$ and voxels with tumor probability greater than 0.5, consistent with the PANTHER whole-pancreas reference. Every voxel in this union is assigned to the most probable pancreatic subregion. Consequently, the head, body and tail masks partition the reported whole pancreas, while tumor remains a separate overlapping mask. OAR masks are obtained by thresholding their region probabilities at 0.5. The anatomical probability hierarchy constrains the subregion and OAR heads, but the independently predicted tumor retains its own spatial extent. All structures are returned by the same network, and the available reference labels determine which are included in a dataset's evaluation. Figure~\ref{fig:method} illustrates the training curriculum and shared prediction pathway.

\subsection{Training and inference}
The supervised training curriculum progressed from whole-pancreas pretraining to subregion pretraining and multitask fine-tuning. The initial whole-pancreas model was trained from random initialization on AMOS and PanSegData for 200 epochs at a learning rate of $10^{-3}$, followed by 50 epochs on the cohort including Cyst-X at $10^{-4}$. Both stages used AdamW and optimized binary pancreas segmentation with Dice and cross-entropy losses. The selected checkpoint initialized a 200-epoch Cyst-X subregion model trained at $10^{-3}$, retaining the encoder-decoder and replacing the binary output heads with a four-class output head for background, body, head and tail. The selected subregion model then initialized the shared encoder-decoder and anatomical output heads of Pancreas-MedNeXt. Tumor and OAR output heads were initialized with zero weights, zero foreground biases and a background bias of two. All parameters were updated during the final 200 epochs of multitask fine-tuning using AdamW at $10^{-3}$, weight decay $3\times10^{-5}$, optimizer epsilon $10^{-4}$ and polynomial learning-rate decay with power 0.9. Each epoch contained 250 optimizer updates and 50 validation-patch iterations, giving 50,000 fine-tuning updates. BF16 activations were combined with FP32 loss calculations. No externally pretrained MedNeXt weights were used, and no encoder or decoder layers were frozen during task transfer.

For training image $i$, let $\mathcal{A}_i$ be the set of structures with available reference masks. Cyst-X contributes whole pancreas and three subregions, PANTHER contributes whole pancreas and tumor, and AMOS contributes whole pancreas and four OARs. Targets are overlapping binary regions. A pancreatic-head voxel, for example, has positive whole-pancreas and head targets, consistent with the probability hierarchy. Missing reference masks exclude their corresponding outputs from the loss, while positive and negative voxels within an annotated region both contribute. This distinguishes a structure that is not labeled from an annotated structure that is absent in a sampled patch. The loss averages binary cross-entropy and soft Dice over the observed regions,
\begin{equation}
\mathcal{L}=\sum_s w_s\frac{1}{B}\sum_{i=1}^{B}\frac{1}{|\mathcal{A}_i|}\sum_{r\in\mathcal{A}_i}
\left[\mathcal{L}_{\mathrm{BCE}}(\hat y_{irs},y_{irs})+
\mathcal{L}_{\mathrm{Dice}}(\hat y_{irs},y_{irs})\right],
\end{equation}
where $s$ indexes decoder supervision scales, $B$ is the batch size, and $\hat y_{irs}$ and $y_{irs}$ are the predicted probability and binary reference for region $r$. Cross-entropy is averaged over voxels, and soft Dice is calculated separately for each image and region with a smoothing constant of $10^{-5}$. Deep-supervision weights are proportional to $[1,1/2,1/4,1/8,0]$ and normalized to sum to one. Normalization by $|\mathcal{A}_i|$ gives each image equal loss weight regardless of the number of annotated structures. Gradients from all observed regions update the shared encoder-decoder, including when only the whole pancreas is labeled and its constituent subregions are unobserved.

Training batches allocated one third of exposure to Cyst-X subregions, one third to AMOS OARs, and one sixth to each PANTHER cohort. This balanced the three segmentation tasks and divided the PANTHER contribution equally between diagnostic and therapeutic MRI, preventing the larger Cyst-X dataset from dominating optimization. Scans were sampled within the assigned task and training partition. Checkpoint selection maximized the exponential moving average of a validation score that equally weighted pancreas DSC, tumor DSC, mean subregion DSC and mean four-OAR DSC. Averaging the subregions and OARs before combining task scores prevented their greater number of channels from dominating selection. Whole-pancreas and subregion pretraining checkpoints were selected separately by validation-patch foreground DSC for their respective tasks. Native-space validation assessed the selected checkpoints rather than providing a second selection criterion. The evaluation thus used one checkpoint per fitted model, selected from validation performance before generating its test predictions. Each PANTHER fold began with the same subregion-pretrained checkpoint, developed without PANTHER images. Multitask fine-tuning and checkpoint selection were repeated for each PANTHER partition while retaining the fixed Cyst-X and AMOS partitions. The first-fold model supplied Cyst-X, AMOS and external MRI predictions.

Each model was trained on one NVIDIA RTX PRO 6000 Blackwell GPU with approximately 96 GB memory using the same fixed random seed. Inference used sliding windows with half-patch steps, Gaussian blending and mirroring, followed by resampling to the original image coordinates. The overlapping windows provided predictions across the full volume, and Gaussian weighting reduced the contribution of patch edges when combining them. One selected checkpoint was used without ensembling. A diagnostic or MR-Linac examination entered through the same single-channel pathway, and the model generated all nine region probabilities without a task-specific network choice. The preprocessing, inference and mask-construction rules were identical across diagnostic and MR-Linac acquisitions, with no scan-specific adaptation at test time.

\subsection{Evaluation and statistical analysis}
Volumetric Dice similarity coefficient (DSC) quantified overlap for each annotated structure. Surface DSC at 3 and 5 mm, the 95th-percentile Hausdorff distance (HD95), mean surface distance and absolute volume error provided complementary geometric measurements \citep{metrics,metricpitfalls}. Surface metrics used surface-area weighting. HD95 was the larger of the two directed weighted 95th-percentile distances, and absolute volume error was calculated from native voxel volumes. The surface tolerances describe agreement at specified spatial scales and are not clinical acceptance thresholds. Pancreas and tumor were evaluated separately in each PANTHER cohort. Head, body and tail were compared with the released Cyst-X subregion labels, and abdominal organs with the AMOS references. Whole-pancreas performance was evaluated wherever its reference was available, allowing gland delineation to be assessed alongside each more specialized task. The external analysis applied the same definitions to the five available organ labels without additional fine-tuning.

Each patient received equal weight in cohort summaries. For Cyst-X, measurements from multiple scans were averaged within each patient before aggregation, preventing additional acquisitions from increasing that patient's contribution. Tables~\ref{tab:results} and \ref{tab:ablation} report DSC as mean $\pm$ sample standard deviation across patients. Tables~\ref{tab:boundaryjoint}--\ref{tab:externalboundary} report mean surface DSC, median HD95 and mean absolute volume error, with mean surface distance additionally reported for PANTHER cross-validation. The standard deviations describe patient variability rather than variability among fold means. Ninety-five percent confidence intervals were obtained from 10,000 percentile bootstrap resamples of patients, preserving the pairing between predictions on the same patient. PANTHER out-of-fold measurements were pooled separately for diagnostic and MR-Linac MRI, rather than averaging folds with different test-set sizes. These intervals quantify patient sampling uncertainty conditional on the fitted models and partitions.

\FloatBarrier
\section{Results}
\subsection{Segmentation performance}
Pancreas-MedNeXt was compared with task-specific nnU-Net, MedNeXt and fine-tuned VISTA3D, and with a unified Swin UNETR model \citep{nnunet2021,mednext,vista3d,swinunetr}. Model configurations are provided in Supplementary Section~\suppref{sec:comparisonmodels}.

\begin{figure}[!htbp]
\centering\includegraphics[width=\linewidth]{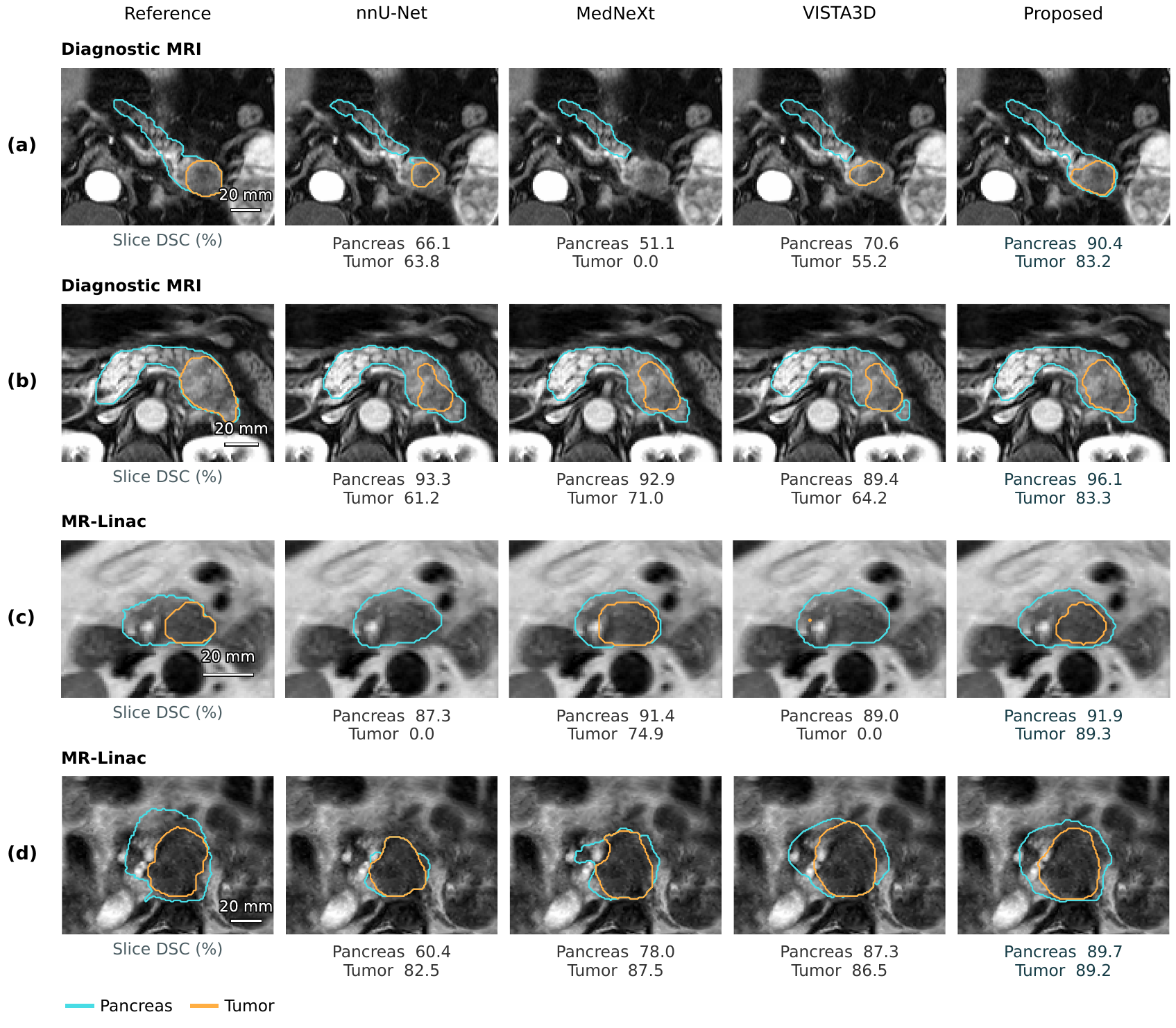}
\caption{Pancreas and tumor contours on diagnostic and MR-Linac MRI. Cyan and amber denote whole pancreas and tumor, respectively. Scores report slice DSCs (\%) for the displayed anatomy.}\label{fig:joint}
\end{figure}

\begin{figure}[!htbp]
\centering\includegraphics[width=\linewidth]{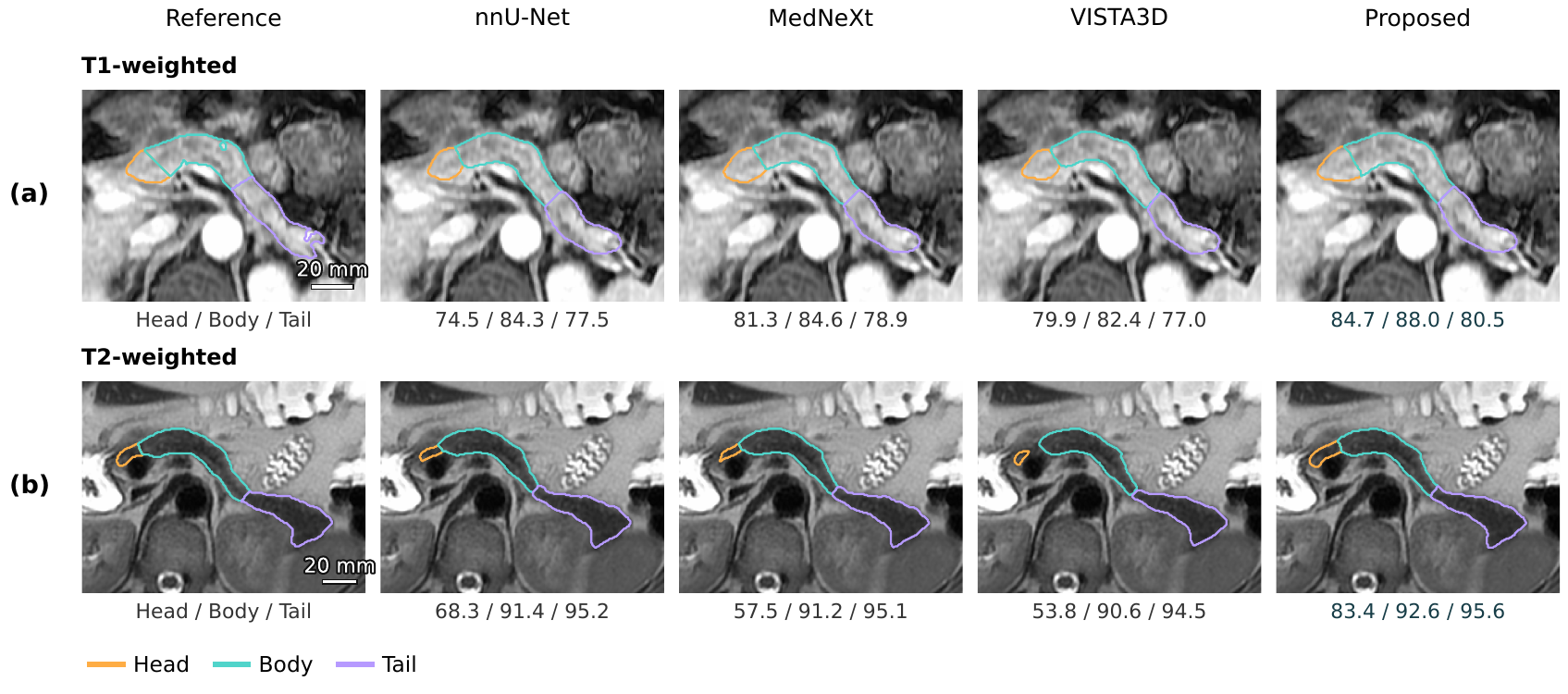}
\caption{Pancreatic subregion contours on T1- and T2-weighted MRI. Score triplets report slice DSC (\%) in head / body / tail order.}\label{fig:parts}
\end{figure}

\begin{figure}[!htbp]
\centering\includegraphics[width=\linewidth]{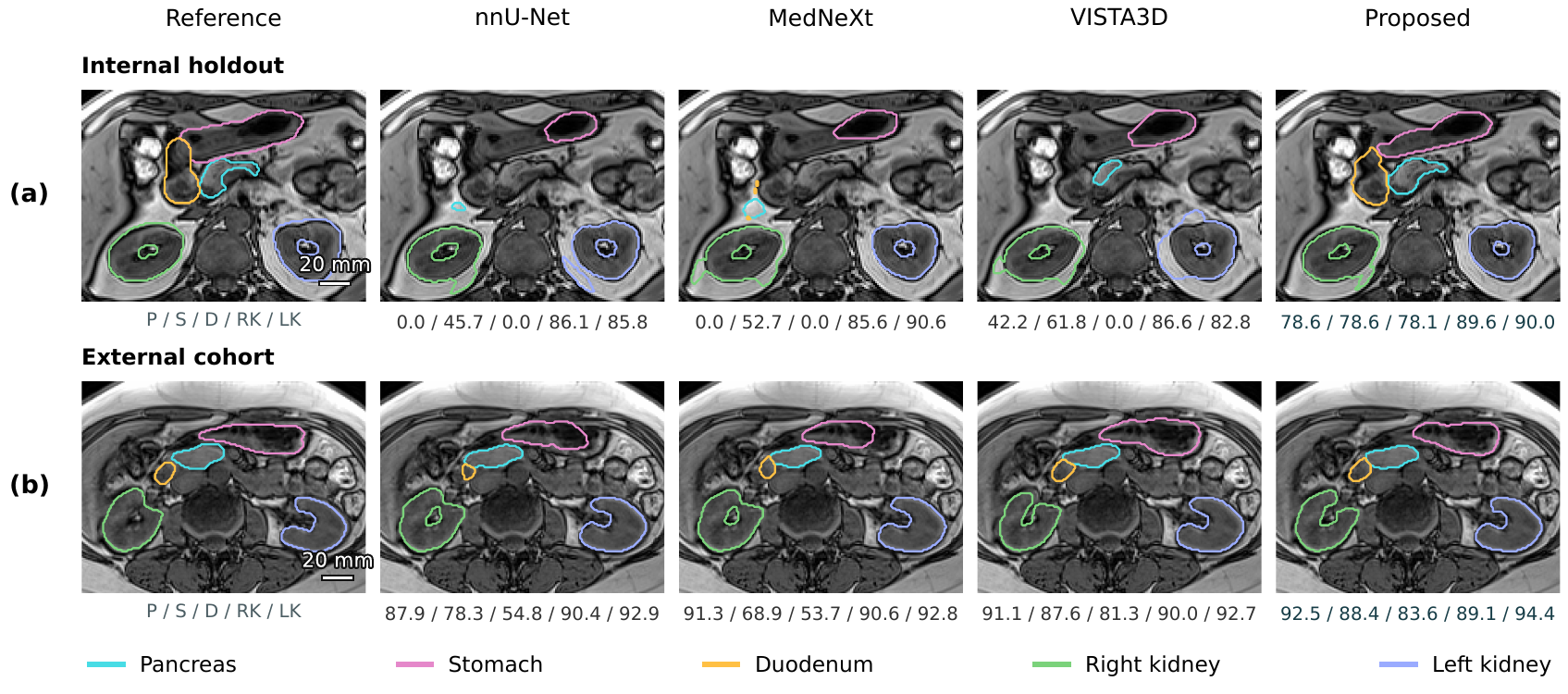}
\caption{Pancreas and abdominal organ contours in (a) the internal AMOS holdout and (b) the external T1-weighted TotalSegmentator MRI cohort. Scores report slice DSCs (\%) in the order P / S / D / RK / LK, denoting whole pancreas, stomach, duodenum, right kidney and left kidney, respectively.}\label{fig:oars}
\end{figure}

In the diagnostic MRI examples, the proposed model follows the reference tumor extent while preserving the adjacent pancreas contour (Figure~\ref{fig:joint}). In the MR-Linac examples, it more closely matches lesion extent and the surrounding gland outline. The Cyst-X examples show head, body and tail segmentation across T1- and T2-weighted appearances. In the T2-weighted example, the most conspicuous difference is recovery of the pancreatic head, while body and tail contours are more similar across methods (Figure~\ref{fig:parts}). In the internal holdout example, the proposed duodenum contour overlaps the reference, whereas the comparison predictions do not (Figure~\ref{fig:oars}). Inspection of the full three-dimensional masks identified small, displaced duodenal predictions from nnU-Net and MedNeXt and an absent prediction from VISTA3D for this patient. In the external T1-weighted example, the proposed model more closely follows the duodenal outline and preserves the adjacent pancreatic contour (Figure~\ref{fig:oars}b). Kidney contours agree more closely across methods than gastrointestinal contours in both examples. The visual differences concern both recovery of an anatomical region and the placement of its boundary, rather than a uniform change in contour size.


\FloatBarrier

In three-fold PANTHER cross-validation, the proposed model achieved the highest mean pancreas and tumor DSC on both diagnostic and MR-Linac MRI among the five evaluated methods (Table~\ref{tab:results}). Compared with task-specific MedNeXt, tumor DSC increased by 4.94 percentage points on diagnostic MRI and 5.89 points on MR-Linac MRI, with paired 95\% confidence intervals of 0.28 to 9.45 and 2.37 to 9.72 points. Whole-pancreas DSC also improved in both cohorts. Pancreas and tumor overlap remained lower on MR-Linac than diagnostic MRI for all methods.

\begin{table}[t]
\caption{Segmentation performance across pancreatic and abdominal tasks. DSC (\%) is reported as mean $\pm$ standard deviation. Bold identifies the highest mean in each row.}\label{tab:results}
\centering\small\setlength{\tabcolsep}{5pt}
\begin{tabular}{@{}llllll@{}}
\toprule
 & \multicolumn{1}{c}{nnU-Net} & \multicolumn{1}{c}{MedNeXt} & \multicolumn{1}{c}{VISTA3D} & \multicolumn{1}{c}{Swin UNETR} & \multicolumn{1}{c@{}}{Proposed} \\
\midrule
\addlinespace[4pt]
\multicolumn{6}{@{}l}{\textit{PANTHER: Diagnostic MRI, three-fold cross-validation}} \\
Pancreas & $80.72\,\pm\,11.98$ & $81.29\,\pm\,12.36$ & $81.81\,\pm\,11.20$ & $76.88\,\pm\,13.54$ & $\boldsymbol{85.73\,\pm\,8.35}$ \\
Tumor & $50.27\,\pm\,29.05$ & $53.47\,\pm\,29.49$ & $52.50\,\pm\,29.93$ & $45.18\,\pm\,29.90$ & $\boldsymbol{58.41\,\pm\,26.53}$ \\
\addlinespace[4pt]
\multicolumn{6}{@{}l}{\textit{PANTHER: MR-Linac MRI, three-fold cross-validation}} \\
Pancreas & $63.76\,\pm\,10.82$ & $64.75\,\pm\,9.87$ & $65.85\,\pm\,9.43$ & $62.82\,\pm\,10.58$ & $\boldsymbol{70.66\,\pm\,11.28}$ \\
Tumor & $31.88\,\pm\,23.20$ & $36.66\,\pm\,25.14$ & $38.46\,\pm\,23.04$ & $29.61\,\pm\,23.90$ & $\boldsymbol{42.55\,\pm\,23.21}$ \\
\addlinespace[4pt]
\multicolumn{6}{@{}l}{\textit{Cyst-X: fixed holdout}} \\
Pancreas & $85.59\,\pm\,6.03$ & $85.56\,\pm\,5.33$ & $82.87\,\pm\,10.16$ & $71.54\,\pm\,12.07$ & $\boldsymbol{86.60\,\pm\,4.40}$ \\
Head & $\boldsymbol{85.85\,\pm\,4.82}$ & $85.41\,\pm\,4.98$ & $84.30\,\pm\,5.32$ & $74.98\,\pm\,7.24$ & $85.57\,\pm\,4.77$ \\
Body & $79.87\,\pm\,7.84$ & $80.29\,\pm\,6.73$ & $76.88\,\pm\,13.42$ & $62.61\,\pm\,15.16$ & $\boldsymbol{80.32\,\pm\,6.33}$ \\
Tail & $77.59\,\pm\,10.80$ & $77.13\,\pm\,11.35$ & $74.81\,\pm\,13.59$ & $57.10\,\pm\,20.12$ & $\boldsymbol{78.82\,\pm\,11.00}$ \\
\addlinespace[4pt]
\multicolumn{6}{@{}l}{\textit{AMOS: fixed holdout}} \\
Pancreas & $78.94\,\pm\,24.65$ & $79.25\,\pm\,21.56$ & $79.57\,\pm\,21.14$ & $80.95\,\pm\,6.02$ & $\boldsymbol{87.37\,\pm\,3.76}$ \\
Stomach & $\boldsymbol{89.47\,\pm\,4.69}$ & $88.68\,\pm\,5.37$ & $86.13\,\pm\,8.40$ & $77.83\,\pm\,14.39$ & $88.03\,\pm\,6.78$ \\
Duodenum & $64.20\,\pm\,25.70$ & $64.51\,\pm\,26.21$ & $65.36\,\pm\,26.79$ & $56.22\,\pm\,22.90$ & $\boldsymbol{69.80\,\pm\,11.15}$ \\
Right kidney & $95.52\,\pm\,5.25$ & $95.32\,\pm\,6.10$ & $94.82\,\pm\,6.86$ & $89.82\,\pm\,14.09$ & $\boldsymbol{95.86\,\pm\,3.68}$ \\
Left kidney & $96.07\,\pm\,3.04$ & $\boldsymbol{96.15\,\pm\,2.80}$ & $95.19\,\pm\,5.16$ & $86.41\,\pm\,20.15$ & $96.00\,\pm\,2.55$ \\
\addlinespace[4pt]
\multicolumn{6}{@{}l}{\textit{TotalSegmentator MRI: external T1-weighted cohort}} \\
Pancreas & $44.05\,\pm\,32.50$ & $40.53\,\pm\,36.55$ & $40.66\,\pm\,33.31$ & $37.83\,\pm\,28.47$ & $\boldsymbol{52.90\,\pm\,30.01}$ \\
Stomach & $50.53\,\pm\,36.59$ & $47.16\,\pm\,38.51$ & $48.22\,\pm\,35.14$ & $41.14\,\pm\,27.34$ & $\boldsymbol{55.45\,\pm\,35.85}$ \\
Duodenum & $31.49\,\pm\,26.74$ & $30.52\,\pm\,27.43$ & $31.44\,\pm\,28.65$ & $21.31\,\pm\,23.30$ & $\boldsymbol{40.23\,\pm\,28.76}$ \\
Right kidney & $57.24\,\pm\,40.85$ & $59.56\,\pm\,40.26$ & $54.95\,\pm\,42.21$ & $42.43\,\pm\,34.08$ & $\boldsymbol{67.74\,\pm\,34.74}$ \\
Left kidney & $59.12\,\pm\,41.11$ & $60.74\,\pm\,42.37$ & $61.72\,\pm\,42.35$ & $41.65\,\pm\,36.01$ & $\boldsymbol{68.52\,\pm\,35.11}$ \\
\bottomrule
\end{tabular}

\end{table}

Across the anatomical tasks, performance depended on the structure. On Cyst-X, the proposed model achieved the highest mean whole-pancreas, body and tail DSC, while nnU-Net obtained a slightly higher head DSC (Table~\ref{tab:results}). On AMOS, the proposed model had the highest mean pancreas, duodenum and right-kidney DSC. Stomach DSC was highest with nnU-Net and left-kidney DSC with MedNeXt. Unified Swin UNETR had lower subregion and gastrointestinal-organ overlap under the evaluated training configuration. The proposed model therefore combined the strongest pancreatic tumor results with favorable performance across regional anatomy and most abdominal organs, while individual task-specific models retained the highest means for the head, stomach and left kidney.

Surface measurements showed greater tumor boundary disagreement on MR-Linac MRI (Table~\ref{tab:boundaryjoint}). For the proposed model, mean tumor surface DSC at 3 mm was 65.42\% on diagnostic MRI and 43.40\% on MR-Linac MRI. Median HD95 was 9.00 and 17.66 mm, respectively. Whole-pancreas surface agreement followed the same pattern. Additional surface and volume measurements for PANTHER, Cyst-X and AMOS are provided in Tables~\ref{tab:boundaryjoint} and \ref{tab:boundaryanatomy}.

\begin{table}[htbp]
\caption{Additional metrics for three-fold PANTHER cross-validation.}\label{tab:boundaryjoint}
\centering\footnotesize\setlength{\tabcolsep}{4pt}
\begin{tabular}{@{}llllll@{}}
\toprule
 & \multicolumn{1}{c}{nnU-Net} & \multicolumn{1}{c}{MedNeXt} & \multicolumn{1}{c}{VISTA3D} & \multicolumn{1}{c}{Swin UNETR} & \multicolumn{1}{c@{}}{Proposed} \\
\midrule
\multicolumn{6}{@{}l}{\textit{Diagnostic MRI, pancreas}} \\
Surface DSC, 3 mm (\%) & 83.35 & 84.34 & 84.70 & 78.72 & \textbf{90.01} \\
Surface DSC, 5 mm (\%) & 90.21 & 90.79 & 91.63 & 87.12 & \textbf{94.95} \\
Median HD95 (mm) & 6.10 (92) & 5.62 (92) & 5.78 (92) & 8.77 (92) & \textbf{3.83} (92) \\
Mean surface distance (mm) & 1.87 (92) & 1.89 (92) & 1.75 (92) & 2.57 (92) & \textbf{1.27} (92) \\
Volume MAE (mL) & 15.46 & 14.45 & 14.28 & 16.70 & \textbf{10.28} \\
\multicolumn{6}{@{}l}{\textit{Diagnostic MRI, tumor}} \\
Surface DSC, 3 mm (\%) & 55.37 & 59.57 & 59.25 & 50.58 & \textbf{65.42} \\
Surface DSC, 5 mm (\%) & 66.26 & 69.62 & 69.56 & 61.84 & \textbf{75.75} \\
Median HD95 (mm) & 9.76 (90) & 9.41 (88) & 9.00 (87) & 12.16 (88) & \textbf{9.00} (90) \\
Mean surface distance (mm) & 6.75 (90) & 5.56 (88) & 5.52 (87) & 9.35 (88) & \textbf{4.72} (90) \\
Volume MAE (mL) & 10.44 & 10.64 & 10.13 & 11.71 & \textbf{9.15} \\
\multicolumn{6}{@{}l}{\textit{MR-Linac MRI, pancreas}} \\
Surface DSC, 3 mm (\%) & 52.40 & 53.76 & 54.36 & 52.58 & \textbf{63.22} \\
Surface DSC, 5 mm (\%) & 68.25 & 69.28 & 70.56 & 68.48 & \textbf{77.58} \\
Median HD95 (mm) & 17.17 (50) & 19.10 (50) & 15.02 (50) & 17.69 (50) & \textbf{13.06} (50) \\
Mean surface distance (mm) & 4.61 (50) & 4.55 (50) & 4.33 (50) & 5.52 (50) & \textbf{3.58} (50) \\
Volume MAE (mL) & 38.62 & 40.51 & 34.21 & 36.99 & \textbf{29.87} \\
\multicolumn{6}{@{}l}{\textit{MR-Linac MRI, tumor}} \\
Surface DSC, 3 mm (\%) & 32.45 & 37.81 & 38.57 & 31.32 & \textbf{43.40} \\
Surface DSC, 5 mm (\%) & 45.47 & 50.61 & 51.73 & 43.60 & \textbf{57.17} \\
Median HD95 (mm) & 19.18 (50) & 18.29 (50) & \textbf{17.22} (48) & 24.90 (49) & 17.66 (50) \\
Mean surface distance (mm) & 7.94 (50) & 7.46 (50) & 6.64 (48) & 8.77 (49) & \textbf{6.51} (50) \\
Volume MAE (mL) & 14.67 & 16.04 & 13.97 & 15.11 & \textbf{13.56} \\
\bottomrule
\end{tabular}

\end{table}

\begin{table}[p]
\caption{Surface agreement, HD95 and volume error for pancreatic subregions and abdominal organs.}\label{tab:boundaryanatomy}
\centering\footnotesize\setlength{\tabcolsep}{4pt}
\textit{Cyst-X}\par\smallskip
\begin{tabular}{@{}llllll@{}}
\toprule
 & \multicolumn{1}{c}{nnU-Net} & \multicolumn{1}{c}{MedNeXt} & \multicolumn{1}{c}{VISTA3D} & \multicolumn{1}{c}{Swin UNETR} & \multicolumn{1}{c@{}}{Proposed} \\
\midrule
\multicolumn{6}{@{}l}{\textit{Pancreas}} \\
Surface DSC, 3 mm (\%) & 91.06 & 91.13 & 89.16 & 74.36 & \textbf{92.13} \\
Surface DSC, 5 mm (\%) & 94.89 & 95.09 & 94.15 & 83.15 & \textbf{95.97} \\
Median HD95 (mm) & 5.38 (43) & 5.05 (43) & 5.44 (43) & 14.14 (43) & \textbf{4.64} (43) \\
Volume MAE (mL) & 7.59 & 8.05 & 11.25 & 14.27 & \textbf{6.80} \\
\multicolumn{6}{@{}l}{\textit{Head}} \\
Surface DSC, 3 mm (\%) & \textbf{90.85} & 90.31 & 89.27 & 76.73 & 90.59 \\
Surface DSC, 5 mm (\%) & \textbf{95.45} & 95.04 & 94.57 & 87.13 & 95.30 \\
Median HD95 (mm) & \textbf{4.55} (43) & 5.16 (43) & 5.36 (43) & 8.52 (43) & 5.40 (43) \\
Volume MAE (mL) & 3.67 & 3.55 & 3.97 & 6.24 & \textbf{3.28} \\
\multicolumn{6}{@{}l}{\textit{Body}} \\
Surface DSC, 3 mm (\%) & 85.46 & 86.08 & 83.41 & 66.82 & \textbf{86.23} \\
Surface DSC, 5 mm (\%) & 91.33 & 92.10 & 90.08 & 76.31 & \textbf{92.45} \\
Median HD95 (mm) & 6.78 (43) & \textbf{6.71} (43) & 7.32 (43) & 15.71 (43) & 7.00 (43) \\
Volume MAE (mL) & 5.69 & 5.73 & 7.31 & 9.30 & \textbf{5.50} \\
\multicolumn{6}{@{}l}{\textit{Tail}} \\
Surface DSC, 3 mm (\%) & 85.00 & 85.54 & 83.38 & 63.14 & \textbf{86.52} \\
Surface DSC, 5 mm (\%) & 90.67 & 91.06 & 90.31 & 71.60 & \textbf{92.13} \\
Median HD95 (mm) & 6.55 (43) & 6.14 (43) & 6.72 (43) & 17.96 (43) & \textbf{6.01} (43) \\
Volume MAE (mL) & 3.00 & 3.40 & 3.10 & 4.81 & \textbf{2.74} \\
\bottomrule
\end{tabular}

\par\medskip\textit{AMOS}\par\smallskip
\begin{tabular}{@{}llllll@{}}
\toprule
 & \multicolumn{1}{c}{nnU-Net} & \multicolumn{1}{c}{MedNeXt} & \multicolumn{1}{c}{VISTA3D} & \multicolumn{1}{c}{Swin UNETR} & \multicolumn{1}{c@{}}{Proposed} \\
\midrule
\multicolumn{6}{@{}l}{\textit{Pancreas}} \\
Surface DSC, 3 mm (\%) & 84.50 & 84.76 & 85.20 & 85.25 & \textbf{93.95} \\
Surface DSC, 5 mm (\%) & 89.47 & 90.24 & 91.19 & 93.01 & \textbf{98.23} \\
Median HD95 (mm) & 4.91 (9) & 4.37 (9) & 4.86 (9) & 6.94 (9) & \textbf{4.00} (9) \\
Volume MAE (mL) & 12.06 & 13.28 & 12.53 & 12.48 & \textbf{5.59} \\
\multicolumn{6}{@{}l}{\textit{Stomach}} \\
Surface DSC, 3 mm (\%) & \textbf{89.43} & 88.50 & 85.00 & 73.05 & 88.32 \\
Surface DSC, 5 mm (\%) & \textbf{94.43} & 93.12 & 91.27 & 81.95 & 93.82 \\
Median HD95 (mm) & \textbf{5.59} (9) & 9.88 (9) & 7.64 (9) & 11.09 (9) & 8.15 (9) \\
Volume MAE (mL) & 17.92 & 19.79 & 20.06 & 19.27 & \textbf{16.87} \\
\multicolumn{6}{@{}l}{\textit{Duodenum}} \\
Surface DSC, 3 mm (\%) & 72.63 & 73.45 & 73.71 & 63.02 & \textbf{79.13} \\
Surface DSC, 5 mm (\%) & 78.16 & 79.59 & 79.87 & 73.13 & \textbf{86.91} \\
Median HD95 (mm) & 20.69 (9) & 10.07 (9) & \textbf{7.29} (8) & 24.00 (9) & 14.23 (9) \\
Volume MAE (mL) & 9.92 & 10.92 & 9.38 & 15.12 & \textbf{5.98} \\
\multicolumn{6}{@{}l}{\textit{Right kidney}} \\
Surface DSC, 3 mm (\%) & 96.49 & 96.10 & 95.46 & 89.33 & \textbf{97.48} \\
Surface DSC, 5 mm (\%) & 98.18 & 97.92 & 97.54 & 92.65 & \textbf{98.85} \\
Median HD95 (mm) & 2.08 (9) & 2.06 (9) & 2.08 (9) & 3.00 (9) & \textbf{2.00} (9) \\
Volume MAE (mL) & 4.03 & 5.00 & 5.57 & 11.57 & \textbf{3.84} \\
\multicolumn{6}{@{}l}{\textit{Left kidney}} \\
Surface DSC, 3 mm (\%) & 97.29 & \textbf{97.92} & 96.14 & 88.50 & 97.82 \\
Surface DSC, 5 mm (\%) & 98.62 & \textbf{99.15} & 97.92 & 93.09 & 99.12 \\
Median HD95 (mm) & 2.50 (9) & 2.78 (9) & \textbf{2.38} (9) & 2.66 (9) & 2.46 (9) \\
Volume MAE (mL) & \textbf{3.59} & 4.54 & 5.46 & 22.44 & 4.95 \\
\bottomrule
\end{tabular}

\end{table}

On external T1-weighted TotalSegmentator MRI, the proposed model achieved the highest mean DSC for all five evaluated organs (Table~\ref{tab:results}). Absolute performance was substantially lower than on AMOS across all methods. Table~\ref{tab:externalboundary} reports additional surface and volume metrics. The external results describe transfer of the locally trained models without further adaptation.

\begin{table}[p]
\caption{External T1-weighted MRI evaluation. Additional metrics use all available organ references within this subset. The coverage analysis reports DSC for scans containing all five reference structures.}\label{tab:externalboundary}
\centering\footnotesize\setlength{\tabcolsep}{3pt}
\textit{Surface agreement and volume error}\par\smallskip
\begin{tabular}{@{}lllllll@{}}
\toprule
 & \multicolumn{1}{c}{nnU-Net} & \multicolumn{1}{c}{MedNeXt} & \multicolumn{1}{c}{VISTA3D} & \multicolumn{1}{c}{Swin UNETR} & \multicolumn{1}{c}{Unified scratch} & \multicolumn{1}{c@{}}{Proposed} \\
\midrule
\multicolumn{7}{@{}l}{\textit{Pancreas}} \\
Surface DSC, 3 mm (\%) & 48.74 & 44.29 & 43.13 & 39.15 & 52.73 & \textbf{57.62} \\
Surface DSC, 5 mm (\%) & 55.34 & 49.20 & 48.81 & 46.32 & 60.00 & \textbf{64.04} \\
Median HD95 (mm) & 40.29 (32) & \textbf{19.42} (25) & 59.95 (31) & 139.88 (33) & 34.62 (33) & 26.86 (31) \\
Volume MAE (mL) & \textbf{11.88} & 15.44 & 18.89 & 45.66 & 26.31 & 26.40 \\
\multicolumn{7}{@{}l}{\textit{Stomach}} \\
Surface DSC, 3 mm (\%) & 49.09 & 46.86 & 45.89 & 34.09 & 47.89 & \textbf{52.58} \\
Surface DSC, 5 mm (\%) & 54.47 & 51.28 & 52.03 & 41.02 & 54.19 & \textbf{58.05} \\
Median HD95 (mm) & 75.61 (32) & 68.40 (30) & 70.28 (33) & 157.65 (33) & 62.14 (32) & \textbf{50.17} (33) \\
Volume MAE (mL) & \textbf{69.47} & 83.19 & 87.55 & 116.64 & 72.91 & 70.67 \\
\multicolumn{7}{@{}l}{\textit{Duodenum}} \\
Surface DSC, 3 mm (\%) & 37.02 & 36.37 & 37.19 & 24.57 & 41.74 & \textbf{46.05} \\
Surface DSC, 5 mm (\%) & 45.01 & 43.71 & 44.74 & 31.45 & 49.84 & \textbf{55.07} \\
Median HD95 (mm) & 120.05 (29) & \textbf{31.46} (27) & 37.08 (25) & 124.43 (31) & 48.45 (31) & 45.28 (31) \\
Volume MAE (mL) & 11.71 & 13.64 & 11.30 & 13.32 & 10.82 & \textbf{8.94} \\
\multicolumn{7}{@{}l}{\textit{Right kidney}} \\
Surface DSC, 3 mm (\%) & 58.58 & 61.27 & 56.66 & 42.46 & 65.94 & \textbf{70.25} \\
Surface DSC, 5 mm (\%) & 61.98 & 64.76 & 60.32 & 48.68 & 70.26 & \textbf{74.06} \\
Median HD95 (mm) & 10.78 (26) & 10.18 (28) & 7.09 (25) & 120.31 (30) & 12.89 (30) & \textbf{5.55} (28) \\
Volume MAE (mL) & 109.84 & 101.84 & 104.64 & 118.51 & 102.64 & \textbf{94.43} \\
\multicolumn{7}{@{}l}{\textit{Left kidney}} \\
Surface DSC, 3 mm (\%) & 60.98 & 62.13 & 62.83 & 42.23 & 65.37 & \textbf{70.55} \\
Surface DSC, 5 mm (\%) & 63.38 & 64.82 & 65.14 & 47.76 & 68.73 & \textbf{74.50} \\
Median HD95 (mm) & 5.71 (28) & \textbf{3.14} (25) & 5.00 (29) & 77.94 (30) & 4.99 (28) & 6.61 (30) \\
Volume MAE (mL) & 81.12 & 81.11 & 81.21 & 98.98 & 80.55 & \textbf{74.30} \\
\bottomrule
\end{tabular}

\par\medskip\textit{Complete-coverage DSC (\%)}\par\smallskip
\begin{tabular}{@{}lllllll@{}}
\toprule
 & \multicolumn{1}{c}{nnU-Net} & \multicolumn{1}{c}{MedNeXt} & \multicolumn{1}{c}{VISTA3D} & \multicolumn{1}{c}{Swin UNETR} & \multicolumn{1}{c}{Unified scratch} & \multicolumn{1}{c@{}}{Proposed} \\
\midrule
Pancreas & 43.28 & 40.51 & 42.80 & 38.31 & 50.01 & \textbf{52.56} \\
Stomach & 51.20 & 48.13 & 45.28 & 37.17 & 50.58 & \textbf{55.17} \\
Duodenum & 29.57 & 27.81 & 30.61 & 20.82 & 31.92 & \textbf{34.40} \\
Right kidney & 56.39 & 60.09 & 56.63 & 40.98 & 63.38 & \textbf{71.35} \\
Left kidney & 59.61 & 61.49 & 60.74 & 38.84 & 63.00 & \textbf{69.95} \\
\bottomrule
\end{tabular}

\end{table}

\subsection{Ablation Studies}
The pretraining strategy was assessed by training an otherwise identical unified model from scratch on the first-fold partition. The comparison therefore evaluates the combined whole-pancreas and subregion pretraining strategy under the same 200-epoch multitask schedule, with the pretrained model additionally receiving the preceding anatomical training stages.

Compared with the same unified model trained from scratch, supervised pretraining improved whole-pancreas DSC in each local cohort and all three pancreatic subregion DSCs (Table~\ref{tab:ablation}). Head, body and tail improvements were 2.01, 3.00 and 3.78 percentage points, respectively, with paired intervals above zero. MR-Linac pancreas, AMOS pancreas and right-kidney gains also had positive intervals. Tumor DSC increased in both imaging cohorts, although these intervals included zero. Stomach and duodenum DSCs were slightly higher after training from scratch, also with intervals spanning zero. On external T1-weighted MRI, pretraining increased mean DSC for all five structures, with positive intervals for the pancreas and right kidney.

\begin{table}[t]
\caption{Effect of supervised pretraining in the unified MedNeXt model. DSC (\%) is mean $\pm$ standard deviation. Differences are proposed minus from scratch, in percentage points with paired 95\% confidence intervals. Bold identifies the higher mean.}\label{tab:ablation}
\centering\small\setlength{\tabcolsep}{6pt}
\begin{tabular}{@{}llll@{}}
\toprule
 & \multicolumn{1}{c}{From scratch} & \multicolumn{1}{c}{Proposed} & \multicolumn{1}{c@{}}{Difference [95\% CI]} \\
\midrule
\addlinespace[4pt]
\multicolumn{4}{@{}l}{\textit{PANTHER: diagnostic MRI, fixed holdout}} \\
Pancreas & $84.35\,\pm\,12.20$ & $\boldsymbol{85.52\,\pm\,10.20}$ & +1.17 [-0.62, +3.82] \\
Tumor & $57.92\,\pm\,25.12$ & $\boldsymbol{59.01\,\pm\,25.87}$ & +1.09 [-3.40, +5.90] \\
\addlinespace[4pt]
\multicolumn{4}{@{}l}{\textit{PANTHER: MR-Linac MRI, fixed holdout}} \\
Pancreas & $72.43\,\pm\,8.45$ & $\boldsymbol{73.55\,\pm\,8.26}$ & +1.12 [+0.16, +2.17] \\
Tumor & $42.30\,\pm\,22.91$ & $\boldsymbol{45.87\,\pm\,22.29}$ & +3.57 [-0.84, +8.52] \\
\addlinespace[4pt]
\multicolumn{4}{@{}l}{\textit{Cyst-X: fixed holdout}} \\
Pancreas & $83.45\,\pm\,7.65$ & $\boldsymbol{86.60\,\pm\,4.40}$ & +3.15 [+2.11, +4.78] \\
Head & $83.56\,\pm\,5.08$ & $\boldsymbol{85.57\,\pm\,4.77}$ & +2.01 [+1.56, +2.50] \\
Body & $77.32\,\pm\,9.54$ & $\boldsymbol{80.32\,\pm\,6.33}$ & +3.00 [+1.54, +5.08] \\
Tail & $75.04\,\pm\,14.11$ & $\boldsymbol{78.82\,\pm\,11.00}$ & +3.78 [+2.61, +5.15] \\
\addlinespace[4pt]
\multicolumn{4}{@{}l}{\textit{AMOS: fixed holdout}} \\
Pancreas & $86.33\,\pm\,4.33$ & $\boldsymbol{87.37\,\pm\,3.76}$ & +1.04 [+0.10, +2.29] \\
Stomach & $\boldsymbol{88.73\,\pm\,5.99}$ & $88.03\,\pm\,6.78$ & -0.69 [-4.44, +2.61] \\
Duodenum & $\boldsymbol{71.53\,\pm\,13.24}$ & $69.80\,\pm\,11.15$ & -1.74 [-5.73, +3.28] \\
Right kidney & $94.88\,\pm\,5.92$ & $\boldsymbol{95.86\,\pm\,3.68}$ & +0.97 [+0.11, +2.54] \\
Left kidney & $95.58\,\pm\,3.66$ & $\boldsymbol{96.00\,\pm\,2.55}$ & +0.43 [-0.08, +1.25] \\
\addlinespace[4pt]
\multicolumn{4}{@{}l}{\textit{TotalSegmentator MRI: external T1-weighted cohort}} \\
Pancreas & $48.29\,\pm\,29.91$ & $\boldsymbol{52.90\,\pm\,30.01}$ & +4.61 [+0.38, +9.57] \\
Stomach & $51.46\,\pm\,34.78$ & $\boldsymbol{55.45\,\pm\,35.85}$ & +3.99 [-2.70, +10.61] \\
Duodenum & $35.38\,\pm\,28.96$ & $\boldsymbol{40.23\,\pm\,28.76}$ & +4.85 [-0.26, +11.53] \\
Right kidney & $61.95\,\pm\,37.82$ & $\boldsymbol{67.74\,\pm\,34.74}$ & +5.79 [+0.06, +12.57] \\
Left kidney & $63.63\,\pm\,39.45$ & $\boldsymbol{68.52\,\pm\,35.11}$ & +4.88 [-1.04, +11.39] \\
\bottomrule
\end{tabular}

\end{table}

\FloatBarrier
\section{Discussion}
Pancreas-MedNeXt enables MRI-based tumor, glandular subregion and abdominal OAR segmentation in a single model and shows strong performance on both MR-Linac and diagnostic MRI. Our approach used partial-label learning and a supervised training curriculum to improve pancreas and tumor overlap over task-specific MedNeXt in diagnostic and MR-Linac cross-validation while retaining regional anatomy and organ segmentation. Unlike a sequence of task-specific networks, the final shared decoder retains subregion segmentation as an active task after tumor and OAR outputs are introduced (Figure~\ref{fig:method}). This extends previous whole-pancreas, tumor and subregion studies toward a combined pancreatic imaging application \citep{pansegnet,liang2020,unifiedpancreas}.

Incomplete annotation coverage is central to the usefulness of this approach. Requiring all structures to be labeled in every scan would substantially restrict the available training population and create a new annotation burden. Instead, the model learns from the labels already supplied for each task while returning a common set of outputs. The hierarchical gland representation is especially relevant because a binary pancreas label and a detailed regional label provide compatible supervision at different anatomical levels. This enables whole-organ and regional datasets to contribute without forcing their annotations into an exclusive label scheme that would discard information. This annotation strategy can further accommodate additional local cohorts that provide only a subset of the existing structures, allowing for easier clinical adoption and finetuning.

The whole-pancreas and abdominal OAR results show that the shared representation supports gland delineation together with surrounding-organ segmentation. Kidney contours agreed closely across methods on AMOS, whereas stomach and duodenum segmentation showed larger differences. The duodenum's curved course around the pancreatic head and proximity to other bowel make localization particularly challenging. In the internal holdout example in Figure~\ref{fig:oars}a, the task-specific predictions were displaced or absent, while the proposed contour recovered the reference location. The external example shows closer agreement with the duodenal boundary alongside preserved pancreatic and gastric contours (Figure~\ref{fig:oars}b). Since the AMOS cohort only contained T1-weighted MRI, we similarly evaluated the external TotalSegmenter dataset only on T1-weighted MRI for fair comparison. The model's organ-level advantage was concentrated in pancreas, duodenum and right-kidney segmentation on AMOS, with the highest mean DSC for all five organs in the external T1-weighted cohort.

TotalSegmentator MRI provided a demanding external benchmark for challenging cases encountered in clinical practice. Sampling from clinical practice, scans included whole-body examinations, pediatric imaging and cases in which their earlier nnUNet-based segmentation model failed. Pediatric examinations introduce smaller organs and different anatomical proportions compared with the adult AMOS cohort used for OAR supervision. Abnormal organ morphology, such as enlarged kidneys, together with whole-body or wide-field-of-view acquisitions extending beyond the abdomen and coarse through-plane resolution, further alters organ appearance and spatial context. These differences may contribute to the lower absolute segmentation accuracy observed across all methods in our external evaluation. Nevertheless, Pancreas-MedNeXt achieved the highest mean DSC for the pancreas and all four OARs with gains of 4.9 to 8.9 percentage points over the next-best method for each organ. We reserved TotalSegmentator MRI for external evaluation to assess robustness under these challenging conditions. The limited size and curated nature of the AMOS training cohort may constrain OAR generalization, and incorporating additional patients with greater acquisition and anatomical diversity could improve performance across broader clinical settings.

Pancreatic subregion segmentation remained effective after the tumor and OAR tasks were introduced. Figure~\ref{fig:parts} shows separation of the head, body and tail across T1- and T2-weighted scans, with more complete recovery of the head in the T2-weighted example. The pretraining ablation (Table~\ref{tab:ablation}) showed that pretraining improved all three subregions, accompanied by higher mean whole-pancreas DSC in both imaging cohorts. The largest regional gain occurred in the tail, where the narrow gland can be difficult to separate from adjacent tissues. Learning gland extent and internal organization before multitask fine-tuning may help preserve these structures as the model acquires new outputs, consistent with the use of anatomical transfer for MRI segmentation \citep{triad_yang}. Beyond whole-organ delineation, the regional masks provide an anatomical basis for quantitative assessment of heterogeneous pancreatic disease, including regional fat analysis \citep{pancreasparts}.

On MR-Linac MRI, surface agreement at margin-scale tolerances gives a first indication of clinical usability. Targets are expanded from the GTV to the CTV and PTV to cover microscopic disease and geometric uncertainty, typically by 3 to 5 mm in MR-guided pancreatic SBRT. Contour deviations smaller than these expansions therefore remain within the planned target volume. Pancreas-MedNeXt achieved the highest tumor surface DSC at 3 and 5 mm (43.40\% and 57.17\%, compared with at most 38.57\% and 51.73\% for the other methods) and the highest pancreas surface DSC (63.22\% and 77.58\%; Table~\ref{tab:boundaryjoint}). OAR agreement is equally important, because the stomach and duodenum are typically the dose-limiting structures in pancreatic SBRT \citep{grimbergen_consensus}. On AMOS, stomach surface DSC at 5 mm was 93.82\%, comparable to the best task-specific model, and duodenal surface DSC was 86.91\%, compared with at most 79.87\% for the other methods (Table~\ref{tab:boundaryanatomy}). Improved delineation of these organs may help refine dose planning where the target abuts the gastrointestinal tract.

Tumor segmentation improved on both diagnostic and MR-Linac MRI, with Pancreas-MedNeXt achieving the highest mean tumor DSC among the evaluated methods (Table~\ref{tab:results}). The improvement over task-specific MedNeXt shows that broader anatomical coverage in a unified model can coexist with stronger lesion delineation. The visual results in Figure~\ref{fig:joint} show closer agreement with lesion extent in both imaging settings, together with improved localization of the surrounding pancreas. These combined outputs place the tumor contour in its glandular context while retaining the adjacent OARs as separate structures for review. All methods had lower pancreas and tumor overlap on MR-Linac than diagnostic MRI, and surface measurements also showed greater tumor contour disagreement in the MR-Linac cohort (Table~\ref{tab:boundaryjoint}). Reduced image quality and tumor conspicuity may partly explain the lower performance, although differences in patient populations and acquisition protocols may also contribute. The gains in both pancreas and tumor segmentation on MR-Linac MRI support the potential use of the model to generate initial contours for clinician review and refinement during MRI-guided adaptive radiotherapy.

\section{Conclusion}
\begin{samepage}
Pancreas-MedNeXt unifies pancreatic tumor, glandular subregion and abdominal OAR segmentation from MRI. Supervised anatomical pretraining improves regional pancreatic segmentation, while the complete multitask approach improves tumor delineation over task-specific MedNeXt on diagnostic and MR-Linac MRI. By providing these structures within one model, the method supports an integrated approach to pancreatic image analysis and contouring for MRI-guided radiotherapy.
\end{samepage}

\section*{Acknowledgments}
This work was supported by National Institutes of Health awards R01HL183223, R01DE033512 and R01CA272991.

\setlength{\bibsep}{3pt}
\bibliographystyle{unsrtnat}
\bibliography{references}
\ifdefined\PMBArticleOnly\else
\clearpage\setcounter{section}{0}
\setcounter{table}{0}
\setcounter{figure}{0}
\renewcommand{\thesection}{S\arabic{section}}
\renewcommand{\thetable}{S\arabic{table}}
\renewcommand{\thefigure}{S\arabic{figure}}
\renewcommand{\theHsection}{S\arabic{section}}
\renewcommand{\theHtable}{S\arabic{table}}
\renewcommand{\theHfigure}{S\arabic{figure}}
\begin{center}
{\Large\bfseries Supplementary material\par}
\medskip
\end{center}

\section{Additional Technical Details}\label{sec:comparisonmodels}
Table~\ref{tab:data} summarizes the datasets and annotated structures. Pretraining and unified training reuse the eligible AMOS and Cyst-X scans. During PANTHER cross-validation, the other two folds each use 80 training, 15 validation and 47 test patients, while Cyst-X and AMOS partitions remain fixed.

Task-specific nnU-Net, MedNeXt and VISTA3D models were trained separately for the PANTHER pancreas/tumor task, Cyst-X subregions and AMOS organs. All methods used the same preprocessing, task partitions, patch size and batch size. nnU-Net and MedNeXt were trained from scratch for 200 epochs using Dice plus cross-entropy loss, with stochastic gradient descent at $10^{-2}$ and AdamW at $10^{-3}$, respectively. Both used polynomial learning-rate decay with power 0.9. VISTA3D adapted the public NV-Segment-CTMR checkpoint through MONAI 1.5.2, with automatic class prompts and no image-specific interaction. Subregion tokens were initialized from the pancreas embedding. The interactive branch remained frozen, and the automatic branch used sigmoid Dice plus binary cross-entropy for 200 epochs with AdamW at a constant $5\times10^{-5}$, weight decay $10^{-5}$ and epsilon $10^{-4}$. 

Unified Swin UNETR used feature size 48, stage depths $(2,2,2,2)$, attention-head counts $(3,6,12,24)$ and activation checkpointing. It was trained from scratch with the proposed model's region outputs, partial-label loss and task sampling. AdamW used a peak rate of $10^{-4}$, weight decay $10^{-5}$, epsilon $10^{-8}$ and ten warmup epochs followed by cosine decay. VISTA3D and Swin UNETR used single-scale losses and internal padding to multiples of 16 and 32, respectively. Each downstream fit used 250 updates and 50 validation-patch iterations per epoch. Checkpoints were selected by validation foreground DSC for task-specific models and the balanced task score for unified models. All five methods completed three-fold PANTHER cross-validation. Cyst-X and AMOS retained fixed task partitions. External evaluation used the selected AMOS or reference-fold unified checkpoint without additional fine-tuning.

Surface DSC is reported as a mean percentage, HD95 as a median distance, and volume MAE as mean absolute volume error. Cyst-X measurements are averaged within each patient before cohort aggregation. Parentheses beside distance metrics identify their contributing cases. For an empty prediction with a nonempty reference, overlap and surface DSC are zero, and undefined distance measurements are omitted. Bold indicates the best value within each row, with lower values favored for distances and volume error.

\FloatBarrier

\fi
\end{document}